# Algorithmic Robustness

*Authors: David Jensen (University of Massachusetts Amherst), Brian LaMacchia (Farcaster Consulting Group, LLC), Ufuk Topcu (University of Texas Austin), Pamela Wisniewski (Vanderbilt University)*

*With Support from: Haley Griffin (Computing Community Consortium)*

*Published: October 17, 2023*

## Introduction

*Algorithmic robustness* refers to the sustained performance of a computational system in the face of change in the nature of the environment in which that system operates or in the task that the system is meant to perform. For example, we might say that an autonomous vehicle's control system is robust if it continues to operate effectively even when it encounters driving conditions that were unforeseen during its initial design and training (e.g., a sudden change in the road, traffic, or weather conditions). Below, we motivate the importance of algorithmic robustness, present a conceptual framework, and highlight the relevant areas of research for which algorithmic robustness is relevant.

Recent concerns about robustness have been particularly acute for systems developed with artificial intelligence (AI) and machine learning (ML) technologies, but such concerns apply more broadly to nearly all computational systems. Complexity, opacity, and brittleness can afflict systems designed and built using traditional methods of software development.

Why robustness? Robustness is an important enabler of other goals that are frequently cited in the context of public policy decisions about computational systems, including trustworthiness, accountability, fairness, and safety. Despite this dependence, it tends to be under-recognized compared to these other concepts. This is unfortunate, because robustness is often more immediately achievable than these other ultimate goals, which can be more subjective and exacting. Thus, we highlight robustness as an important goal for researchers, engineers, regulators, and policymakers when considering the design, implementation, and deployment of computational systems.

We urge researchers and practitioners to elevate the attention paid to robustness when designing and evaluating computational systems. For many key systems, the immediate question after any demonstration of high performance should be: "How robust is that performance to realistic changes in the task or environment?" Greater robustness will set the stage for systems that are more trustworthy, accountable, fair, and safe.

Toward that end, this document provides a brief roadmap to some of the concepts and existing research around the idea of algorithmic robustness.

**Motivation**

Concerns about algorithmic robustness have grown as computational systems have become more central in societal decision making. Algorithmic approaches have been applied to a wide variety of tasks previously performed entirely by humans, including medical diagnosis, loan approval, parole decisions, employment screening, piloting aircraft, and driving automobiles. A variety of social science findings indicate that human operators often place inappropriate trust in automated systems, even when those systems are explicitly designed to augment, rather than replace, human decision making.

Concerns about algorithmic robustness have also grown as computational systems have simultaneously become more complex, capable, and opaque. Some of the most salient examples include self-driving automobiles, large language models such as ChatGPT, and facial recognition systems. In particular, experience with these systems indicates that human operators can find it extremely difficult to forecast what changes in task and environment might negatively affect the performance of these computational systems.

Finally, concerns have grown as systems have come to rely on machine learning models. Statistics and machine learning technologies have long been used to produce systems whose performance can exceed that of humans on certain limited tasks. While such increased performance is often accompanied by unexpected brittleness, human supervision has served as an effective fail-safe. On the other hand, recent advances in machine learning have often decreased, rather than increased, the degree to which human observers can understand the basis of the systems' output and forecast whether such output will be affected by changing conditions.

In particular, the increasing use of deep learning technologies has produced end-to-end systems with remarkable competence in typical situations. However, the performance of these systems can drop precipitously when seemingly incidental changes to the environment or task are introduced.[1] These systems can be characterized as "not knowing what they know" — they can be "confidently wrong" and thus mislead human operators. The result is that systems based on modern machine learning can have sharp performance "cliffs" in which performance degrades catastrophically with little warning or predictability.

---

[1] In machine learning, these changes are sometimes referred to as *distribution shift*, *concept shift*, or *dataset shift*. Recent work has also described the resulting problem as *out-of-distribution* (OOD) inference.

**Conceptual framework**

To better understand what is meant by "algorithmic robustness," it is useful to decompose any given application into its *system*, *task*, and *environment*. In a typical application, a *system* (a.k.a., algorithm, agent, piece of software), performs a *task* in an *environment*. For example, an autonomous vehicle (the system) is given a specific destination (the task) by a user, and then the system attempts to navigate to that destination given the available roads, traffic conditions, and weather (the environment).

The performance of a given system is typically measured in terms of one or more desirable behaviors of the system (e.g., time, distance, fuel usage, safety, etc.) with respect to the task in a given environment. *Robustness* is a property of systems that minimize performance changes when aspects of the task or environment change.[2,3] For example, we might say that a particular autonomous vehicle is robust if its safety and navigation efficacy are relatively unchanged even though a user alters their desired destination (a change in the task) or major new construction projects are started in the region surrounding those destinations (a change in the environment).[4]

On its face, the algorithmic robustness of a given system might appear relatively easy to evaluate empirically. Researchers could run experiments that change various aspects of the task or environment, and then determine whether key performance measures of the system change. However, robustness is typically a *relative* rather than an *absolute* concept. Changing aspects of the task or environment can make the task inherently more or less difficult, such that even a theoretically optimal system will perform better or worse given the change. For example, in the autonomous vehicle example above, the inherent task difficulty might be affected by weather, traffic, amount of road construction, destination, time of day, and many other factors. This also implies that robustness is a *continuous* rather than a *binary* characteristic: Systems are not robust or non-robust. Instead, they are *more* or *less* robust than alternative systems.[5]

Another important consideration is the *approach* used to produce robustness. Systems can be robust because they are:

- *Invariant* — The system is unresponsive to changes to the task or environment. For example, an autonomous vehicle might drive in the same manner regardless of lighting, temperature, or road conditions. Such behavior might produce robustness with respect to performance indicators that measure delivery time.

---

[2] This parallels other definitions such as the one provided by ISO ("ability of a system to maintain its level of performance under a variety of circumstances"( ISO/IEC TS 5723:2022) and the related discussion in NIST's *AI Risk Management Framework 1.0*.
[3] Clearly, some performance changes are unavoidable (e.g., the minimum time required to reach a goal will change with the distance of that goal from a start state). However, robust systems generally limit the performance changes in response to environment or task changes. See the next paragraph for additional discussion.
[4] In this document, we explicitly focus on changes to the task and environment that are *non-adversarial*. That is, the changes are not explicitly designed by an adversary to change system performance, but occur more naturally or incidentally. System response to adversarial change is a very important topic, and is more commonly encompassed under the rubrics of "security" or "resilience".
[5] Despite this, this document will sometimes refer to systems being "robust." This should be read as implying that they are "more, rather than less, robust."

- *Responsive* — The system responds appropriately to the environment, and thus is robust. No information is carried over between different task instances (i.e., no learning), so the same behavior will happen if the same environment is encountered again. For example, an autonomous vehicle might drive more slowly when the road is obscured by snow, leaves, or other substances (e.g., wind-driven dust). Such behavior might produce robustness with respect to performance indicators that measure safety.
- *Adaptive* — The system itself changes in response to the environment, resulting in greater robustness. Information is carried over between task instances, so a more responsive behavior will occur if the same environment is encountered again. For example, an autonomous vehicle might learn from experience how wind-driven dust affects braking time and adapt accordingly. Such behavior might produce robustness with respect to performance indicators such as long-term trade-offs between safety and performance.

Note that these approaches are not mutually exclusive, but instead can be used in combination. A single system might employ all three approaches, depending on the task and environmental changes that it encounters. For example, a system might be designed and constructed so that it is *invariant* under one set of potential changes, but then become *responsive* (by trying each of a set of fixed alternative strategies) or *adaptive* (by learning new strategies) if it detects other sorts of changes to its task or environment.

All other things being equal, *invariant* systems are typically preferred over *responsive* or *adaptive* systems, because *invariant* systems require no additional time to adapt. However, systems that are successfully *responsive* and *adaptive* can be expected to sustain performance over a wider range of potential changes to the task or environment.

**Relevant research areas**

Unsurprisingly, a variety of research efforts within the computer science community have aimed to advance the state-of-the-art in algorithmic robustness, often supported by future-looking federal research programs. For example, the National Science Foundation (NSF) and the Defense Advanced Research Projects Agency (DARPA), two of the largest supporters of federal research programs in computing, each support multiple efforts related to algorithmic robustness. These include:

From NSF:

- *Secure and Trustworthy Cyberspace* — The Secure and Trustworthy Cyberspace (SaTC) program[6] is an interdisciplinary and cross-cutting initiative charged with the goal of protecting and preserving the growing social and economic benefits of cyber systems while ensuring security and privacy by finding fundamentally new ways to design, build, and operate cyber systems; protect existing infrastructure; and motivate and educate individuals about cybersecurity.

---

[6] https://new.nsf.gov/funding/opportunities/secure-trustworthy-cyberspace-satc

- *Cyber-Physical Systems* — The Cyber-Physical Systems (CPS) program[7] supports core research needed to engineer complex cyber-physical systems, some of which may also require dependable, high-confidence, or provable behaviors. Functionality enabled by learning and artificial intelligence is gaining importance in cyber-physical systems rapidly. As cyber-physical systems become more data-rich, the potential to move toward autonomous designs is growing and robustness and safety are increasingly central.

- *Safe Learning-Enabled Systems* — The Directorate for Computer and Information Science and Engineering (CISE) released a new solicitation[8] in 2023 focused on Safe Learning-Enabled Systems which focuses on foundational research that leads to the design and implementation of learning-enabled systems in which safety and robustness is ensured with high levels of confidence.

From DARPA:

- *Competence-Aware Machine Learning* — A first step toward robustness is knowledge of the conditions under which a given system's performance will change. Recent research has sought to characterize system *competence*, particularly when those systems are based on machine learning. DARPA's Competency-Aware Machine Learning (CAML) program aims to enable learning systems to be aware of their own competence and thus "know what they know".

- *Explainable Artificial Intelligence* — A key feature of trustworthy AI systems is the ability to explain the reasoning of such systems. Without such explanations, human operators won't know when and to what degree the inferences of such systems can be trusted. Recent research has sought to develop methods to explain otherwise opaque AI systems and to enable development of systems whose inferences are inherently more explainable than those of modern machine learning models. DARPA's recent Explainable AI (XAI) program aimed to create machine learning techniques that could produce more explainable models, while maintaining high performance, and enable human users to understand, appropriately trust, and effectively manage modern AI systems.

- *Domain adaptation* and *transfer learning* — If models are not directly transportable, they can often be repaired or adapted to new environments and tasks by minimal retraining. This approach is often referred to as *domain adaptation* or *transfer learning*. This area was the focus of a DARPA program more than a decade ago.[9]

- *Adaptation to open-world novelty* — Research in computational systems often assumes a "closed" world in which the potential environmental conditions are entirely known. Recent work has sought to develop systems that act appropriately and effectively in novel situations that occur in *open worlds*. DARPA's Science of Artificial Intelligence and Learning for Open-world Novelty (SAIL-ON) program aims to develop the scientific

---

[7] https://new.nsf.gov/funding/opportunities/cyber-physical-systems-cps
[8] https://www.nsf.gov/pubs/2023/nsf23562/nsf23562.htm
[9] Senator, T. E. (2011). Transfer learning progress and potential. *AI Magazine*, 32(1), 84-84.

- principles necessary to detect, characterize, and adapt to novelty in open-world domains.

- *Adversarial learning* — Changes in tasks and environments can sometimes be made intentionally to disrupt the performance of a computational system. These changes can even be made in ways that exploit the fact that a given system is based on machine learning (e.g., by influencing training data or the learning algorithm itself). For example, DARPA's program in Guaranteeing AI Robustness Against Deception (GARD) aims to establish theoretical foundations for identifying system vulnerabilities, characterizing properties that will enhance system robustness, and encouraging the creation of effective defenses.

- *Life-long learning* — Many systems based on machine learning cease learning after an initial training phase, but other systems have been designed to continue learning after deployment. Some of these latter systems exhibit troubling behavior when tasks or environments shift, "forgetting" previously learned knowledge and skills as they learn new ones. This reduces robustness when tasks or environments revert to earlier patterns. Researchers have sought to reduce or eliminate such "forgetting".  For example, DARPA's recent Lifelong Learning Machines (LLM) program aimed to develop systems that can learn continuously during execution and become increasingly expert while performing tasks, are subject to safety limits, and apply previous skills and knowledge to new situations, all without forgetting what has been previously learned.

In conclusion, algorithmic robustness is an indispensable feature for computational systems, which increasingly affect virtually every aspect of our lives. It is important that we strive to develop new computational systems that not only are trustworthy, accountable, fair, and safe but also sustain these desirable features under reasonable changes in their tasks and the environment in which they operate. It is also necessary to establish new research programs that aim to evaluate the algorithmic robustness of computational systems and equip future computational systems with algorithmic robustness.